\documentclass[letterpaper, 10 pt, conference]{ieeeconf}  

\IEEEoverridecommandlockouts                              

\usepackage{graphics} 
\usepackage{epsfig} 
\usepackage{mathptmx} 
\usepackage{times} 
\usepackage{amsmath} 
\usepackage{amssymb}  
\usepackage{url}
\usepackage[utf8x]{inputenc} 

\title{\LARGE \bf
	3D tracking of water hazards with polarized stereo cameras
}

\author{Chuong V. Nguyen$^{1}$, Michael Milford$^{2}$ and Robert Mahony$^{1}$
	\thanks{$^{1}$Chuong Nguyen and Robert Mahony are with the Australian Centre for Robotic Vision, Research School of Engineering, Australian National University, Canberra ACT 2601, Australia
		{\tt\small Firstname.Lastname@anu.edu.au}}%
	\thanks{$^{2}$Michael Milford is with Australian Centre for Robotic Vision, School of Electrical Engineering and Computer Science, Queensland University of Technology, Brisbane QLD 4000, Australia
		{\tt\small michael.milford@qut.edu.au}}%
}

\begin{document}

	\maketitle
	\thispagestyle{empty}
	\pagestyle{empty}

	\begin{abstract}
		
		Current self-driving car systems operate well in sunny weather but struggle in adverse conditions.
		One of the most commonly encountered adverse conditions involves water on the road caused by rain, sleet, melting snow or flooding.
		While some advances have been made in using conventional RGB camera and LIDAR technology for detecting water hazards, other sources of information such as polarization offer a promising and potentially superior approach to this problem in terms of performance and cost.
		In this paper, we present a novel stereo-polarization system for detecting and tracking water hazards based on polarization and color variation of reflected light, with consideration of the effect of polarized light from sky as function of reflection and azimuth angles.
		To evaluate this system, we present a new  large `water on road' datasets spanning approximately 2 km of driving in various on-road and off-road conditions and demonstrate for the first time reliable water detection and tracking over a wide range of realistic car driving water conditions using polarized vision as the primary sensing modality. Our system successfully detects water hazards up to more than 100m.
		Finally, we discuss several interesting challenges and propose future research directions for further improving robust autonomous car perception in hazardous wet conditions using polarization sensors.
		
	\end{abstract}

	\section{INTRODUCTION}
	
	Polarization of light in outdoor environments is very common.
	Scattered light from the sky is polarized \cite{illguth2015effect}, especially close to the horizon.
	Light reflected from water surfaces \cite{wehner2001polarization} is also polarized, particularly when the angle of reflection is low.
	Some vegetation and many man-made objects, especially those with glass, polarize light \cite{vanderbilt1985polarization, walraven1981polarization}.
	Water on the road poses a significant hazard when driving a vehicle, so the utility of polarized light for detecting it has obvious appeal, especially in the age of self-driving cars where perception systems must match current human capability.
	
	Several authors have considered using polarization as cue to identify water surfaces.
	Xie et al \cite{xie2007polarization} used three cameras with polarizers to estimate polarizing angle and detect water surfaces.
	Stereo cameras equipped with horizontal and vertical polarizers have been used to detect still and running water bodies \cite{yan2014water} or wet ground \cite{kim2016wet}.
	While Yan \cite{yan2014water} used hand-tuned parameters for classification, Kim et al \cite{kim2016wet} used Gaussian Mixture Model (GMM) for hypothesis generation and a Support Vector Machine (SVM) for hypothesis verification.
	For puddle detection, Kim et al \cite{kim2016wet} used RANSAC to perform ground plane depth fitting and then employed a threshold technique on depth information to detect puddles.
	%
	%
	
	Detection of water hazards from a single camera is also possible using only color and texture information.
	Zhao et al \cite{zhao2013research} developed SVM-based classification of color and texture information from a single image.
	Rankin and Matthies \cite{rankin2010daytime} proposed a technique based on variation in color from sky reflections as a camera moves closer to a puddle. Later Rank et al \cite{rankin2011daytime} proposed another technique that searches for water reflection matching with the sky. Rankin's techniques work well for wide open space, well defined water bodies and distance greater than 7m.
	
\begin{figure}
		\centering
		\includegraphics[width=\columnwidth]{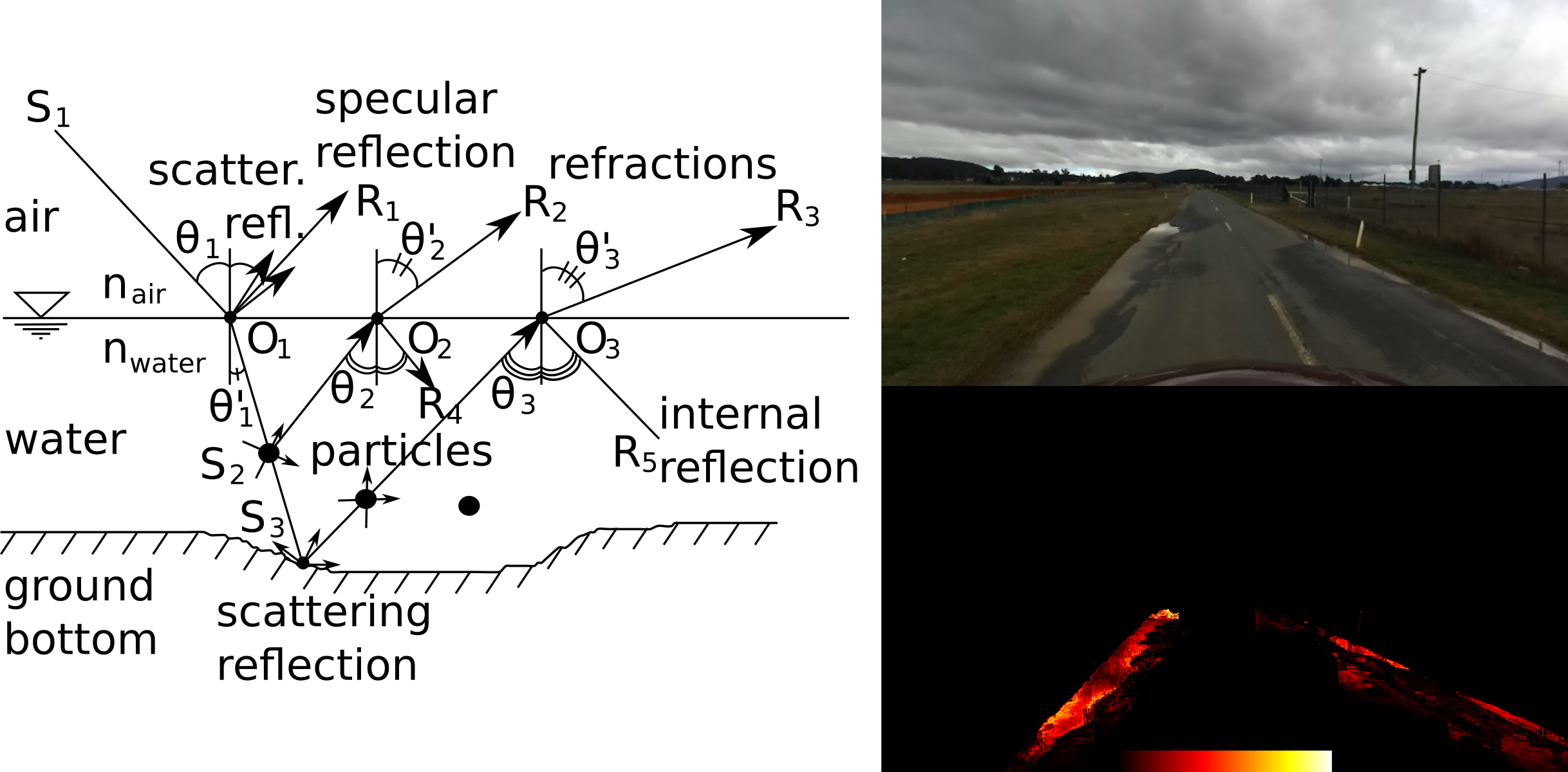}
		\caption{Model of polarized sky light interaction with water surface for detection of water hazards.}
		\label{fig:front_figure}
	\end{figure}

	In this paper we use polarization effect as the primary cue to identify water in driving conditions. To improve on past research, we take a novel approach using the change in the color saturation and polarization as functions of reflection and azimuth angles to account for effect from sky polarization. The method is illustrated in Fig.~\ref{fig:front_figure}.
	
	\section{Light polarization and reflection on water surfaces}
	\label{sec:polarization_reflection}

	In this section, we provide a treatment of the theoretical background behind the use of polarization to detect water surfaces. We review the Rayleigh sky model of polarization as described in \cite{illguth2015effect} and a reflection model proposed by  \cite{rankin2010daytime}. To provide the theoretical background for the novel approach in this paper, we describe how incident light polarization, scattering and refraction can be used to model the color and the polarization of light received from water surfaces.
	
	\subsection{Review of existing work}
	
	Ambient light is often treated as non-polarized; however this is not true for light coming from sky. Illguth \cite{illguth2015effect} shows that sky polarization variations during the day can affect the appearance of building glass fa{\c{c}}ades, which presumably also applies to water surfaces. Polarized light is generated by Rayleigh scattering effect.
	Following the Rayleigh-sky model \cite{illguth2015effect}, the degree or intensity of the polarized light component is a function of scattering angle $\gamma$:
	
	\begin{equation}
	\eta = \frac{\eta_{max} \sin^2 \gamma}{1 + \cos^2 \gamma}
	\label{eq:rayleigh_scattering}
	\end{equation}
	
	\begin{equation}
	\cos \gamma = \sin \theta_{sun} \sin \theta_{view} \cos \psi + \cos \theta_{sun} \cos \theta_{view}
	\label{eq:scattering_angle}
	\end{equation}
	
	\noindent where $\theta_{view}$ is  viewing angle between a viewed point on sky and zenith, $\theta_{sun}$ angle between the sun and zenith, and $\psi$ azimuth angle between the sun and the viewed point on the sky.
	
	\begin{figure}[thpb]
		\centering
		\includegraphics[width=\columnwidth]{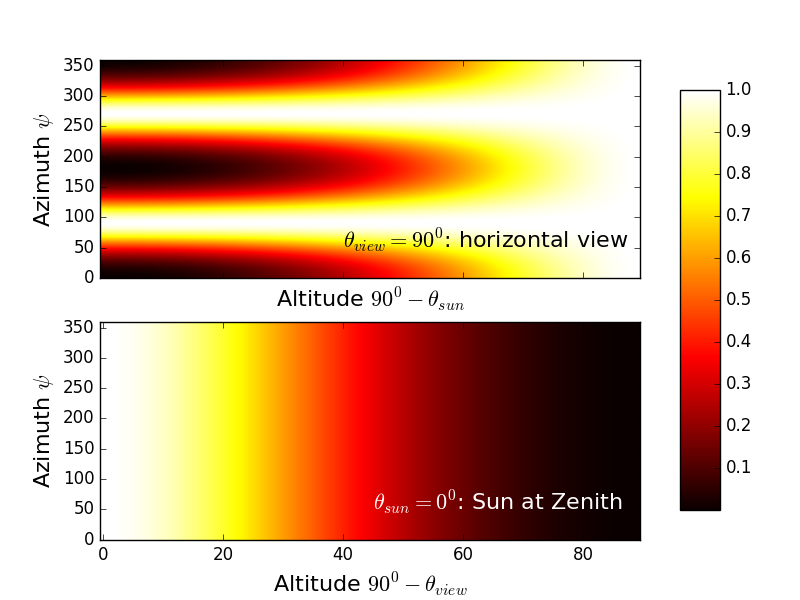}
		\caption{Degree or intensity of sky light polarization as function of viewing angle $\theta_{view}$ (between a viewed point on sky and zenith) and angle of the sun $\theta_{sun}$ (between the sun and zenith).}
		\label{fig:sky_polarisation}
	\end{figure}
	
Fig.~\ref{fig:sky_polarisation} visualizes equations \ref{eq:rayleigh_scattering} and \ref{eq:scattering_angle}. 
The top of Fig.~\ref{fig:sky_polarisation} shows that when observing the horizon ($\theta_{view}=90$), the degree of polarization is maximal at azimuth angles of 90 and 270 degrees from the sun when it is near the horizon ($\theta_{sun} ~= 90^0$), and maximal across the entire horizon when the sun is at its zenith ($\theta_{sun} = 0^0$). 
The bottom of Fig.~\ref{fig:sky_polarisation} shows that when the sun is at zenith ($\theta_{sun} = 0$), the degree of polarization is maximum at the ground horizon. In reality, the maximum degree of polarization can reach up to 90\% of the total light energy.
	
Although sky light is generally polarized, existing research often avoids dealing with this effect \cite{rankin2010daytime, yan2014water, kim2016wet}. With an assumption of unpolarized incident light, Rankin et al. \cite{rankin2010daytime} modeled total water reflection as a summation of specular reflection from the water surface $R_{ref}$, scattering reflection of water surface $R_{scatter}$, scattering reflection of particles in water $R_{particles}$, and scattering reflection from the ground bottom of the water $R_{bottom}$:
	
	\begin{equation}
	R_{total} = R_{reflect} + R_{scatter} + R_{particles} + R_{bottom}
	\label{eq:total_reflection}
	\end{equation}
	
	Specular reflection on water is known to polarizes light \cite{wehner2001polarization, xie2007polarization, rankin2010daytime, kim2016wet}. Specular reflection from the water surface $R_{reflect}$ is a sum of two polarization components $R_{reflect,\bot}$ and $R_{reflect,\parallel}$, perpendicular and parallel respectively to the plane formed by the incident and reflected rays:
	
	\begin{equation}
	R_{reflect,\bot}(n_1, n_2, \theta) = \left[ \frac{n_1 \cos \theta - n_2 \sqrt{1 - (n_1 / n_2)^2  \sin^2 \theta}}{n_1 \cos \theta + n_2 \sqrt{1 - (n_1 / n_2)^2  \sin^2 \theta}} \right]^2
	\label{eq:refl_coef_perp}
	\end{equation}
	
	\begin{equation}
	R_{reflect,\parallel}(n_1, n_2, \theta) = \left[ \frac{n_1 \sqrt{1 - (n_1 / n_2)^2  \sin^2 \theta} - n_2 \cos \theta}{n_1 \sqrt{1 - (n_1 / n_2)^2  \sin^2 \theta} + n_2 \cos \theta} \right]^2
	\label{eq:refl_coef_paral}
	\end{equation}
	
\noindent where $n_1$ and $n_2$ are refractive indices of first and second media (here air and water, respectively) and $\theta$ is reflection angle at the medium interface.

The higher the coefficient $R_{reflect,\bot}$, the stronger the intensity and apparent color of the sky as it appears in the reflection on water surface.
The difference between the two coefficients $R_{reflect,\bot}$ and $R_{reflect,\parallel}$ gives the polarization in the reflected light. However, Equations \ref{eq:refl_coef_perp} and \ref{eq:refl_coef_paral} alone do not explain the influence of sky polarization on the polarization, the color and intensity of reflection on water surfaces in reality.
	
\subsection{Effect of polarized sky light on polarization of reflection}
	
\begin{figure}
		\centering
		\includegraphics[width=0.6\columnwidth]{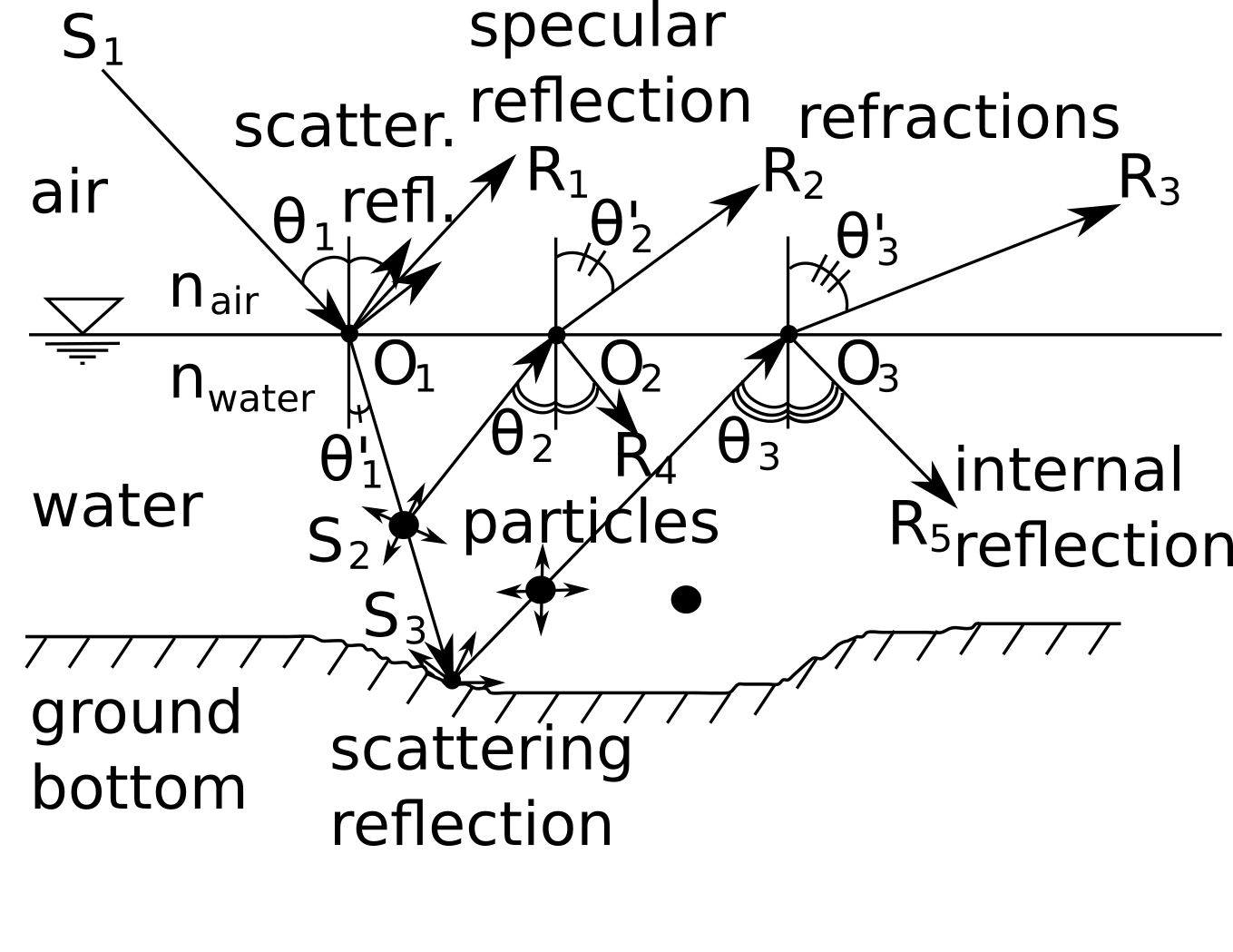}
		\caption{Light reflection and refraction in a water puddle with dirt particles and ground bottom.}
		\label{fig:reflection_puddle}
	\end{figure}

	To take into account the effect of polarization of sky light on water reflection, we propose a model of light interaction with water. Fig. \ref{fig:reflection_puddle} illustrates the 4 components of equation \ref{eq:total_reflection}. The scattering reflection component due to surface roughness at $O_1$ is very small compared to specular reflection for water, unless there is a large concentration of particles or it is windy. In addition, the light component exiting the water from particles $R_{particles}$ and ground bottom $R_{bottom}$ are in fact refractions, not reflections. As a result, the term "water reflection" as sensed by eyes or cameras should be understood as a sum of reflection and refraction.
	
	Internal reflection coefficients within water at $O_2$ and $O_3$ are also given by equations \ref{eq:refl_coef_perp} and \ref{eq:refl_coef_paral} with $n_1 = n_{water}$ and $n_2 = n_{air}$. Refraction coefficients (from air to water or from water to air) can be obtained from reflection coefficients:
	\begin{equation}
	R_{refract,\bot}(n_1, n_2, \theta') = 1 - R_{reflect,\bot}(n_1, n_2, \theta)
	\label{eq:refr_coef_perp}
	\end{equation}
	
	\begin{equation}
	R_{refract,\parallel}(n_1, n_2, \theta') = 1- R_{reflect,\parallel}(n_1, n_2, \theta)
	\label{eq:refr_coef_paral}
	\end{equation}
	
	\noindent where Snell's law gives:
	\begin{equation}
	\sin \theta' = \frac{n_1}{n_2} \sin \theta
	\label{eq:snell_law}
	\end{equation}
	
	
	Suppose that polarized light from the sky comprises energy components $E_{\bot}^S(\theta, \psi)$ and $E_{\parallel}^S(\theta, \psi)$ for perpendicular and parallel components respectively as functions of reflection angle and azimuth angle. Light that enters water surface illuminates particles and the ground bottom. The total energy entering the water is therefore:
	
	\begin{multline}
	F^S = E_{\bot}^S(\theta, \psi) [1- R_{reflect,\bot}(n_1, n_2, \theta)] \\
	 +  E_{\parallel}^S(\theta, \psi) [1- R_{reflect,\parallel}(n_1, n_2, \theta)]
	\end{multline}
	
	Part of the energy $F^S$ is scattered by suspended particles and ground bottom while the rest is absorbed:
	
	\begin{equation}
	\mu_{particles} + \mu_{bottom} + \mu_{absorption} = 1
	\end{equation}
	where $\mu_{particles}$ and $\mu_{bottom}$ are scattering coefficients for particles and the ground bottom,  and $\mu_{absorption}$ is the absorption coefficient for both particles and the ground.
	
	Due to random scattering and internal reflection process, light within water can be considered highly unpolarized (the perpendicular component is approximately the same as parallel component). Part of the scattered light comes out of the water via refraction. The total light energy component coming out of water is the summation of reflection and refraction (viewed at the same point and angle) for each polarization component:
	
	\begin{multline}
	E_{\bot}^R(\theta, \psi) \approx E_{\bot}^S(\theta, \psi) R_{reflect,\bot}(n_1, n_2, \theta) + \\
	0.5 F^S [\mu_{particles} + \mu_{bottom}] R_{refract,\bot}(n_1, n_2, \theta'=\theta)
	\label{eq:reflr_energy_perpend}
	\end{multline}
	
	\begin{multline}
	E_{\parallel}^R(\theta, \psi) \approx E_{\parallel}^S(\theta, \psi) R_{reflect,\parallel}(n_1, n_2, \theta) + \\
	0.5 F^S [\mu_{particles} + \mu_{bottom}] R_{refract,\parallel}(n_1, n_2, \theta'=\theta)
	\label{eq:reflr_energy_parallel}
	\end{multline}
	
	
	\begin{figure}[thpb]
		\centering
		\includegraphics[width=\columnwidth]{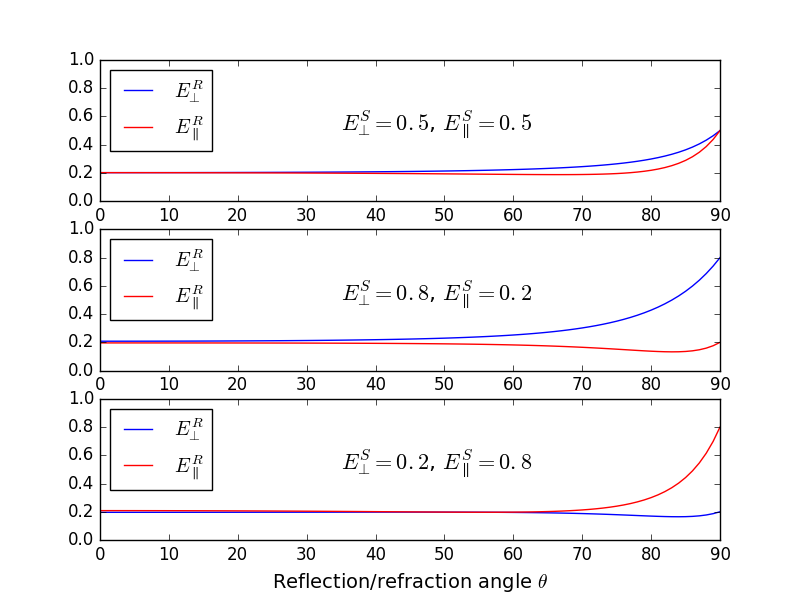}
		\caption{Reflection and refraction from water as function of degree and direction of polarization. Top: unpolarized light source. Middle: light source that is 80\% polarized in perpendicular direction with reflection plane. Bottom: light source that is 80\% polarized in parallel direction with reflection plane.}
		\label{fig:reflection_polarisation}
	\end{figure}
	
	Equations \ref{eq:reflr_energy_perpend} and \ref{eq:reflr_energy_parallel} are visualized in Fig~\ref{fig:reflection_polarisation}. For illustrative purpose, the absorption coefficient $\mu_{absorption}$ is set to 60\% here. Depending on the position of the sun and the viewed point on the sky, a different ratio of $E_{\bot}^S(\theta, \psi)$ and $E_{\parallel}^S(\theta, \psi)$ is obtained. Fig~\ref{fig:reflection_polarisation} shows that for unpolarized light (top) and polarized light perpendicular to reflection plane (middle), $E_{\bot}^R(\theta, \psi)$ is higher than $E_{\parallel}^R(\theta, \psi)$. This agrees with the conventional assumption \cite{yan2014water, kim2016wet}. However for polarized light of 80\% in parallel direction to the reflection plane (bottom of Fig~\ref{fig:reflection_polarisation}), $E_{\bot}^R(\theta, \psi)$ is lower than $E_{\parallel}^R(\theta, \psi)$. This is however different from conventional assumption.
	Fig.~\ref{fig:reflection_polarisation} also shows that at reflection angles above 70 degrees (or at a large distance), the polarization difference between perpendicular and parallel components is large and thus provides a strong detection cue for water hazards at a large distance.
	
	Equations \ref{eq:reflr_energy_perpend} and \ref{eq:reflr_energy_parallel} are significant as they explain the mechanism of the color mixing process for the sky, suspended particles and water bottom. As $\theta$ and $\psi$ vary with viewpoint, the polarized components of light coming from different sources vary accordingly, leading to color variation on the water surface. Furthermore, when capturing an image of the water surface with orthogonal polarizers, the orthogonal components of polarized light can be captured separately and show up as changes in the image color and intensity dependent on the direction of the filters. We use these phenomena as the basis for our water hazard detection process
		
	\section{Approach}
	
	With the theoretical foundation provided in the previous section, we now describe our approach to visual detection of water hazards. The overall algorithm is as follows:	
	
	\begin{enumerate}
		\item Calculate disparity map from a stereo image pair.
		\item Estimate ground plane disparity by 3D plane fitting using robust estimation within a triangular region in front of the car. Obtain the equation of the horizon line.
		\item Use ground plane disparity to warp the right-hand image to the left-hand image, producing point correspondence for feature extraction.
		\item Compute the reflection and azimuth angles for all pixels below horizon line.
		\item Extraction image color features from point correspondences as function of reflection and azimuth angle.
		\item Training: manually created ground truth masks are used to train Gaussian Mixture Models for water and not-water.
		\item Testing: trained Gaussian Mixture Models are used to compute likelihoods of water and not-water pixels, producing a water mask for each stereo image pair.
	\end{enumerate}
	
	\subsection{Stereo disparity and ground-plane disparity estimation}
	Given a pair of stereo images, a variant of semi-global block matching algorithm \cite{hirschmuller2008stereo, birchfield1998pixel} is employed to obtain a disparity map.
	
\begin{figure}
		\centering
		\includegraphics[width=\columnwidth]{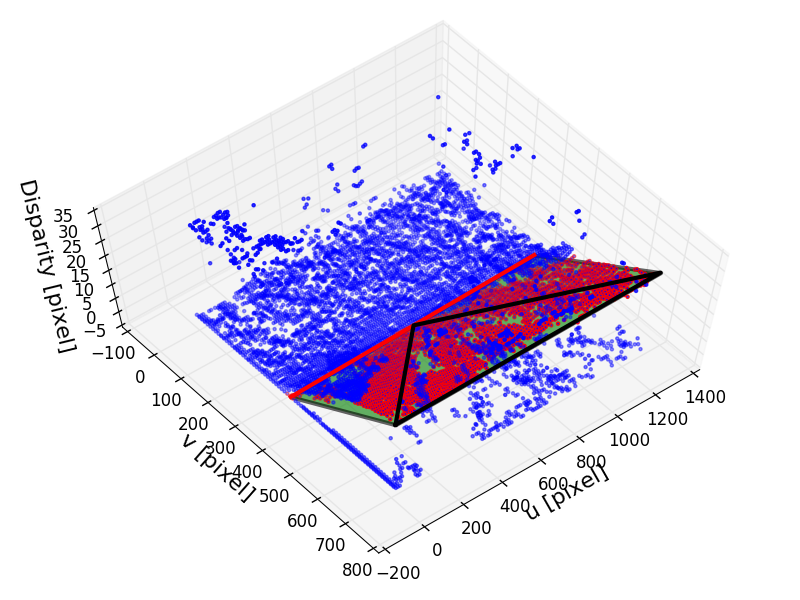}
		\caption{Plane fitting to stereo disparity map. Data points within the black triangle are selected for fitting. Red points are inliers and blue points are outliers when the fitted plane extends to the whole image. The red line represents the horizon line where disparity is zero.}
		\label{fig:plane_fitting}
	\end{figure}

	Plane fitting by robust estimation with Cauchy loss function is applied to extract ground disparity shown as black rectangle in Fig. \ref{fig:plane_fitting}. Horizon line is the zero disparity line on the fitted ground  disparity plane show as red line in Fig.~\ref{fig:plane_fitting}. If the 3D plane equation of ground disparity is $au + bv + c = \delta$ where $\delta$ is disparity and $u, \; v$ are pixel coordinates, the horizon line equation is $au + bv + c = 0$.
	
	
	For simplicity, assuming there are no obstacles on the ground near the car, the right image can be warped to the left image using the fitted ground disparity. Direct pixel-to-pixel comparison between the left image and warped right image is now possible.
	
	
	\subsection{Reflection and azimuth angles}
	
\begin{figure}
		\centering
		\includegraphics[width=0.9\columnwidth]{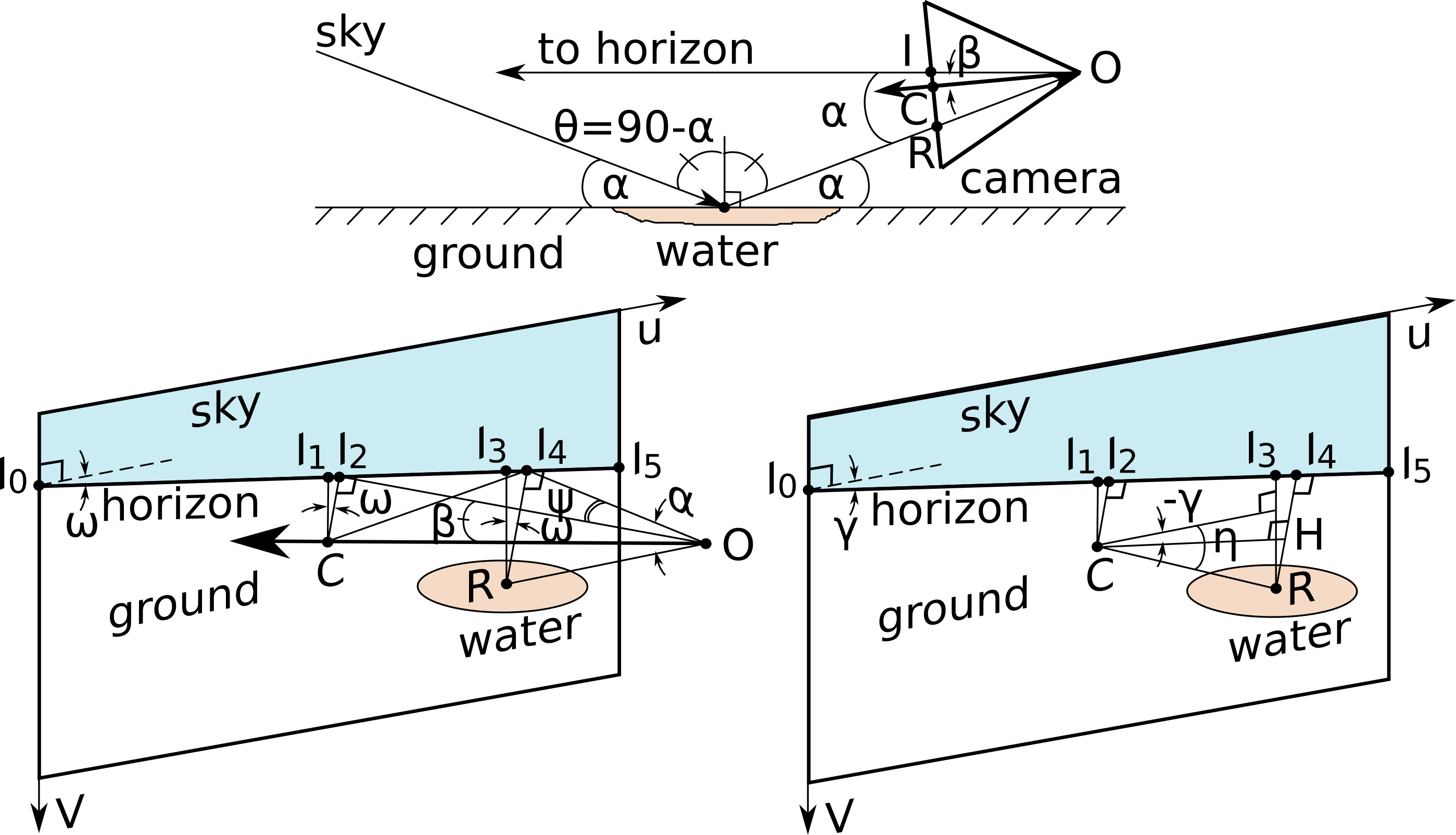}
		\caption{Top: schematic side view of reference camera (left) and the scene when the horizon is completely horizontal to the camera image. Bottom: geometry of camera and reflection when the horizon line is not horizontal. O is camera origin, C camera optical center, I and $I_0$ to $I_5$ points on horizon line image, R image of reflection point on water or ground, $\theta$ incident/reflection angle, $\psi$ azimuth angle, $\beta$ and $\omega$ relative angles of horizon line image to optical axis and image horizontal direction.}
		\label{fig:reflection_angle}
	\end{figure}

	As discussed in section \ref{sec:polarization_reflection}, the reflection angle is one of the main factors affecting the color and polarization of refection from water. Here, we propose an algorithm to compute reflection angle from the pixel position and horizon line as illustrated in Fig. \ref{fig:reflection_angle}. Here we assume that water surface is parallel to horizon line, and that camera optical axis $OC$ is perpendicular to the image plane.
	
	Reflection angle $\theta = (\pi/2-\alpha)$, where $\alpha = \widehat{ROI_4}$ ($RI_4$ is perpendicular to horizon line) can be obtained by cosine rule:
	\begin{equation}
	\alpha = \arccos \left(\frac{OR^2 + OI_4^2 - RI_4^2}{2*OR*OI_4} \right)
	\end{equation}
	
	\noindent where:
	\begin{align}
	\begin{split}
	RI_4 &= RI_3 \cos \omega = (v_R - v_{I_3}) \cos \omega
	\label{eq:RI4}
	\end{split}\\
	\begin{split}
	OR &= \sqrt{OC^2 + CR^2} = \sqrt{f^2 + (u_C-u_R)^2 + (v_C - v_R)^2}
	\label{eq:OR}
	\end{split}\\
	\begin{split}
	OI_4 &= \sqrt{OC^2 + CI_4^2} = \sqrt{f^2 + (u_C - u_{I_4})^2 + (v_C - v_{I_4})^2}
	\label{eq:OI4}
	\end{split}\\
	\begin{split}
    u_{I_4} &= u_R + RI_4 \sin \omega = u_R + (v_R - v_{I_3}) \cos \omega \sin \omega
	\label{eq:uI4}
	\end{split}\\
	\begin{split}
	\omega &= \arctan \left(\frac{v_{I_5} - v_{I_0}}{u_{I_5} - u_{I_0}} \right)
	\label{eq:gamma}
	\end{split}
	\end{align}
	\indent and $f$ is focal length.
	
	To account for sky polarization effect, azimuth angle $\psi$ is also computed at each pixel position. For simplicity, the azimuth angle is obtained with respect to camera forward direction for the tests in this paper (without taking account camera orientation in the earth coordinate system), as the camera mostly points along a single orientation and sky polarization in opposite directions are similar enough:

	\begin{align}
	\begin{split}
	\psi &= \arctan \left( \frac{I_2I_4}{OI_2} \right)= \arctan \left( \frac{CR \; \cos(\eta - \omega)}{\sqrt{OC^2 + CI_2^2}} \right)
	\label{eq:phi}
	\end{split}\\
	\begin{split}
    u_{I_2} &= u_C + CI_2 \sin \omega = u_C + (v_C - v_{I_1}) \cos \omega \sin \omega
	\end{split}
	\label{eq:uI2}
	\end{align}
	\noindent where $\eta$ is angle of CR formed with pixel horizontal direction in the image.
	
	\subsection{Gaussian Mixture Model for water-ground classification}
	As discussed in section \ref{sec:polarization_reflection}, color and polarization (i.e. intensity) vary with incident and azimuth angles. A Gaussian Mixture Model (GMM) is used to capture these effects:
	
	\begin{equation}
	p(x; a_k; S_k; 	\pi_k) = \Sigma_{k=1}^m \pi_k p_k(x),\: \pi_k > 0,\: \Sigma_{k=1}^m \pi_k = 1
	\end{equation}
	\begin{equation}
	p(x) = \frac{1}{\sqrt{(2\pi)^d |S_k|} } \exp \left( -\frac{1}{2}(x-a_k)^\intercal S_k^{-1} (x-a_k)  \right)
	\end{equation}
	where $\pi_k$ is the weight of the k-th Gaussian cluster, $p_k$ is a normal distribution with mean $a_k$ and covariance $S_k$.
	
	Images are converted from RGB to HSV color space and features at pixels below horizon line were collected for training and classification of the GMM models. After many validations, we found that strong features for detection are saturations (left and right cameras), value or brightness (left camera), reflection and azimuth angles (left camera). Saturations and value are normalized to vary from 0 to 1.
	
	\section{Experimental setup}
	
\begin{figure*}
		\centering
		\includegraphics[width=\textwidth]{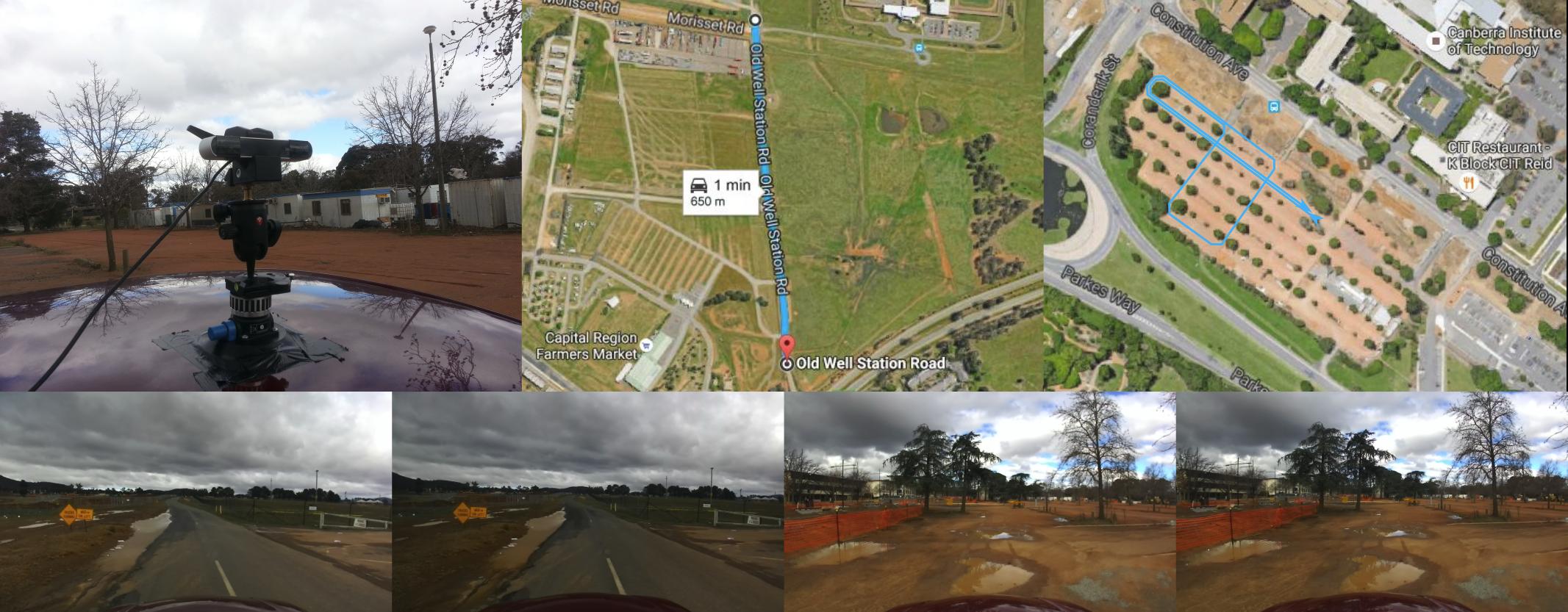}
		\caption{Experimental setup. Top left: ZED stereo camera on top of car with polarizing films on camera lenses. Top middle and right: trajectories of on-road and off-road video sequences. Bottom row: examples of stereo image pairs from on-road and off-road video sequences.}
		\label{fig:camera_setup}
	\end{figure*}

	Images were captured using a laptop-connected StereoLabs ZED stereo camera \cite{Stereolabs2016} mounted on top of a car at the height of 1.77m as shown in Fig.~\ref{fig:camera_setup}. Each stereo video frame is a side-by-side left-right image of $(2\times1280)\times720$ pixel resolution. The nominal camera focal length is 720 pixels with a stereo baseline of 120mm. Linear polarizing films \cite{3dlens2016} with max transmittance of 42\% and polarizing efficiency of 99.9\% are attached to the front of camera lenses in the horizontal direction (left view) and vertical direction (right view). A Python script acquired images from the stereo camera at 30 frames per second, added them to a buffer queue which was streamed to the solid-state drive of the laptop to avoid frame dropping. Stereo calibration was done using StereoLabs calibration software. Stereo reconstruction was performed using semi-global block matching algorithm implemented by the OpenCV library \cite{bradski2008learning}.
	
	Two video sequences of real driving conditions (Fig.~\ref{fig:camera_setup}) were selected to test our algorithm:
	\begin{itemize}
		\item The on-road sequence contains 5357 video frames comprising a 1360m round trip of country road (location between 35°13'19.6"S 149°09'09.9"E and 35°13'40.9"S 149°09'12.1"E).
		\item The off-road sequence contains 6098 frames comprising a 780m round trip in a car park (location at 35°17'13.1"S 149°08'09.0"E) next to a roadwork site.
	\end{itemize}
	
	For each sequence, ground-truth masks were created manually from images containing a number of water puddles of various sizes and distances. Training images were selected such that they always contained water puddles of various sizes and distances. Test images were selected at approximately even intervals throughout the video sequences, including frames that contain no puddles. These datasets and ground-truth masks are made freely available at \cite{Nguyen2016_water_puddle_data_url}.
	
	\section{Results}
	
	
	
	For each sequence, masks and corresponding stereo images were used to train two GMM models of 5 Gaussian clusters, one for water and one for not-water. Numbers of ground-truth water masks for training and testing the GMM models are given in Table \ref{tab:validation}.
	
\begin{figure*}[p]
		\centering
		\includegraphics[width=\textwidth]{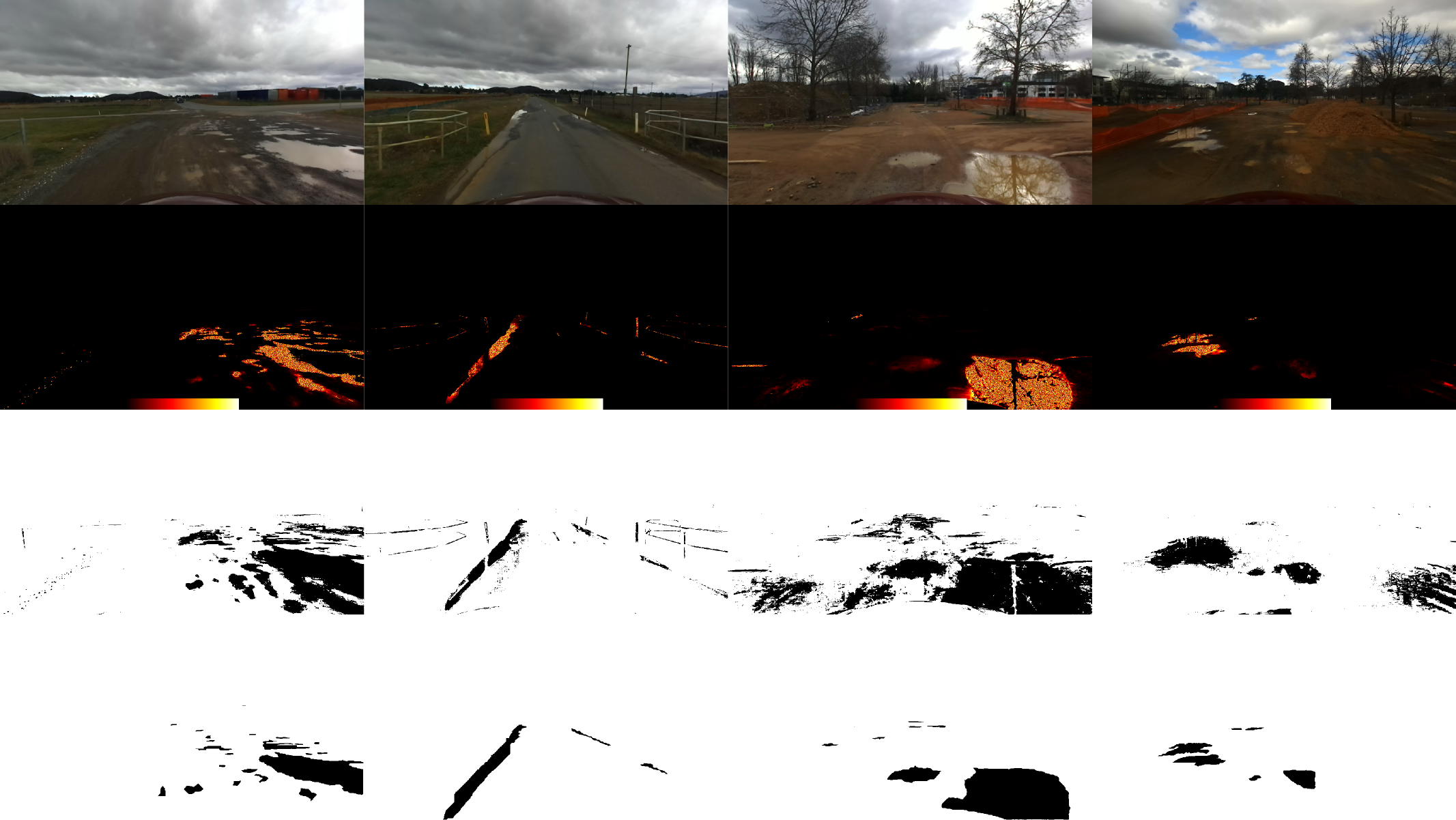}
		\caption{Example snapshots of water detection for the on road (left columns) and off road (right columns) sequences. From top to bottom rows are stereo left images, predicted ratio of water and not-water likelihoods, water masks (where the ratio is higher than 1), and ground truth water masks. Water areas show strong ratio value while some wet areas also show high ratio value. The color bar ranges from 0 to 255.}
		\label{fig:prediction_on_off_road}
	\end{figure*}

	For testing, the trained GMM models were used to compute the likelihood of pixels being water and not-water from features extracted from individual stereo image pair. The ratio of the likelihood of water over that of not-water shown in the second rows of Fig.~\ref{fig:prediction_on_off_road} is used to determined if a pixel belongs to water or not. The third and fourth row of 	
	Fig.~\ref{fig:prediction_on_off_road} show examples of prediction masks, and ground truth masks. The black regions of prediction masks indicate water pixels where the ratio is larger than 1. These examples were produced by the GMM models with azimuth angle.
	
	True positives (real water), true negatives (real not-water), false positives (wrong water), and false negatives (wrong not-water) were obtained by comparing pixels between predicted masks and ground-truth masks. Accuracy, recall and precision performance were then computed. Table \ref{tab:validation} provides the details of training, testing and the prediction results using GMM models with left-right saturations, left brightness, reflection angle, and without and with azimuth angle. This is to test the effectiveness of taking account the effect of polarization from the sky.
	
	\begin{table}
		\caption{Validation of the water detection algorithm without and with azimuth angle feature.}
		\label{tab:validation}
		\begin{tabular}{l*{4}{c}r}
			Seq. & Train & Test & Accuracy & Recall & Precision\\
			\hline
			On-road & 54 & 65 & 0.956/0.970 & 0.906/0.860 & 0.247/0.378\\
			Off-road & 71 & 80 & 0.933/0.931 & 0.824/0.837 & 0.223/0.220
		\end{tabular}
	\end{table}	
	
	Test accuracy and recall are excellent, above 90\% and 80\% respectively, for both sequences and both GMM models. The test precision is however relatively low, at 25\% and 22\% for on-road and off road respectively, using our GMM models without azimuth angle therefore ignoring the effect of sky polarization. With our GMM model including azimuth angle, the precision of on-road sequence increases to 38\% while that of off-road sequence stays the same. This shows that our theory of sky polarization affecting the appearance of water applies well to the on-road sequence where most water puddles reflect light directly from the sky. However, this theory does not apply successfully for the off-road case where there are trees blocking light from the sky and altering the relationship.
	
	\begin{figure}
		\centering
		\includegraphics[width=\columnwidth]{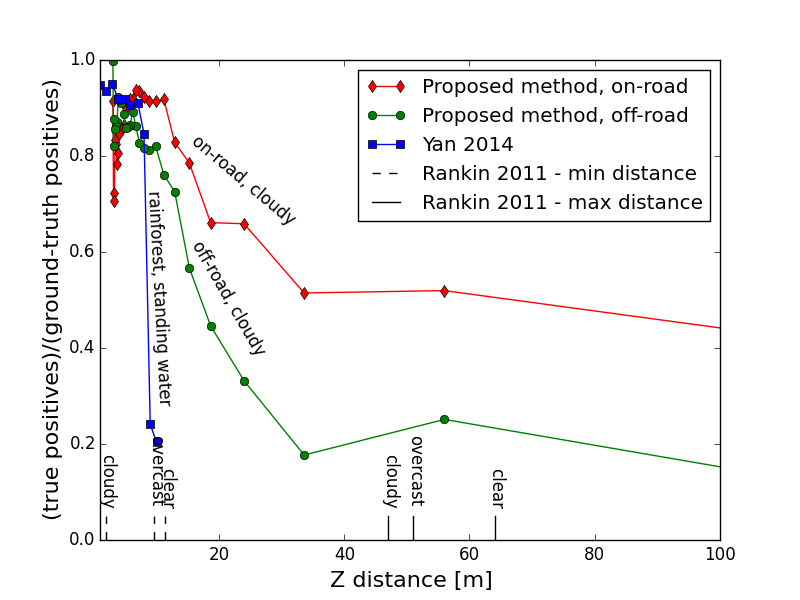}
		\caption{Water detection range for on-road and off-road sequences together with results by Yan \cite{yan2014water} and Rankin et al \cite{rankin2011daytime}.}
		\label{fig:detection_range}
	\end{figure}
	
	For both sequences, there are several source of error. Objects protruding above the ideal ground plane lead to some false detections. Such false detections can be mitigated by excluding above-ground trees and obstacles. Wet ground also reflects and polarizes light. As a result, wet areas are often classified as water puddles, although the ground-truth masks don't include them, leading to reduced  precision performance. For hazard detection, it is however useful to detect wet ground. Furthermore to some extend the ground-truth masks have errors as human can miss out some small water regions where GMM models can pick up correctly. The precision is also affected in this case. Finally, images captured by the ZED camera in this work often has noisy stripes in dark image areas. These noisy stripes slightly change image brightness and perceived polarization characteristics therefore adversely affecting the detection algorithm.
	
	Fig. \ref{fig:detection_range} shows effective detection range using our GMM model with azimuth angle to account for sky polarization effect. Vertical axis is ratio between water true positives and water ground truth positives. For both on-road and off-road sequences, the true detection rate is around 90\% for distance from 3 to 10m and then decreases gradually for distance between 10 to 35m and then stays almost the same for distance from 35m to about 60m (considering the size of the water puddles are quite small at this distance range). For off-road sequence, the true detection is lower than for on-road sequence, presumably due to reflection of trees. True detection drops to 20\% at distance of about 35m. As a crude comparison, results achieved in previous studies by \cite{rankin2010daytime} and \cite{yan2014water} are also included,  although these results were obtained from different datasets (not publicly available) with slightly different conditions. Yan \cite{yan2014water} used a similar polarized stereo camera system and achieved about 90\% true detection rate for images captured in rain forest up to 8m distance then drop down sharply to 20\% at 10m distance and above. Rankin and Matthies \cite{rankin2011daytime} did not provide statistics of detection at different distances but max and min distances when water was firstly detected in cloudy, overcast and clear sky conditions with a single large water body.
	
	Technically the system is correctly detecting (small) amounts of water on the ground at a wide range of distance, but for practical applications it is likely we would need to find a way to discount non-significant water coverage, perhaps by attempting to calculate the depth of the water hazard from the changes of the color of the water as function of viewing angle. Noisy water masks could be cleaned up by filtering across multiple image frames.
	
	\section{CONCLUSIONS}
	
	Water on the ground poses a fundamental hazard to robots and manned or autonomous vehicles. While there has been significant research on this problem to date, most has involved using conventional sensing modalities or have been limited studies using light polarization. In this paper, we have attempted to advance the concept of using light polarization to detect water hazards on the ground in a self-driving car context. In particular we have developed a novel approach that uses both polarization and color variation of reflected light to detect water. Effect of sky polarization is accounted by the use of azimuth angle which significantly improves detection precision for on-road sequence. From this project, we have developed two new datasets for on-road and off-road driving in wet conditions which we provide to the community, and demonstrated water detection on these datasets using the proposed approach. We hope that the research presented here provides a new stimulus for further investigations of the utility of using conventional camera technology equipped with polarizing filters for improving robot and autonomous vehicle perception in the myriad of environmental conditions where water poses a hazard.

	\addtolength{\textheight}{-12cm}
	
	\section*{ACKNOWLEDGEMENT}
	
	This research was conducted by the Australian Research Council Centre of Excellence for Robotic Vision (CE140100016) \url{http://www.roboticvision.org}.
The authors would like to thank Prof Richard Hartley of Research School of Engineering, Australian National University, for his kind support to this project, and Mr Alexander Martin of the Research School of Engineering who helped record early water-on-road videos.

	\bibliographystyle{IEEEtran}
	\bibliography{IEEEabrv,water_detection}

\end{document}